\documentclass[sigconf]{acmart}
\AtBeginDocument{%
  }

\setcopyright{acmlicensed}
\copyrightyear{2018}
\acmYear{2018}
\acmDOI{XXXXXXX.XXXXXXX}
\acmConference[Conference acronym 'XX]{Make sure to enter the correct
  conference title from your rights confirmation email}{June 03--05,
  2018}{Woodstock, NY}
\acmISBN{978-1-4503-XXXX-X/2018/06}

\usepackage{fontawesome5}

\usepackage{booktabs}   
\usepackage{adjustbox}  
\usepackage{multirow}   
\usepackage{caption}    

\begin{document}


\newcommand{\rebus}{|\faIcon{sync-alt} \faIcon{bus}|}
\newcommand{\rebusprog}{\textsc{RebusProg}}
\newcommand{\rebusdescprog}{\textsc{RebusDescProg}}
\newcommand{\rebusdescprogice}{\textsc{RebusDescProgICE}}

\title[\rebus: Rebus Puzzles Understanding Benchmark]{\rebus: A Large and Diverse Multimodal Benchmark for evaluating the ability of Vision-Language Models to understand Rebus Puzzles}


\author{Trishanu Das}
\authornote{Both authors contributed equally to this research.}
\email{dastrishanu01@gmail.com}
\affiliation{%
  \institution{Tredence Inc.}
  \country{India}
}

\author{Abhilash Nandy}
\authornotemark[1]
\email{nandyabhilash@gmail.com}
\affiliation{%
  \institution{Indian Institute of Technology Kharagpur}
  \country{India}
}

\author{Khush Bajaj}
\affiliation{%
  \institution{Indian Institute of Technology Kharagpur}
  \country{India}
}

\author{Deepiha S}
\affiliation{%
  \institution{Indian Institute of Technology Kharagpur}
  \country{India}
}

\begin{abstract}
  Understanding Rebus Puzzles (Rebus Puzzles use pictures, symbols, and letters to represent words or phrases creatively) requires a variety of skills such as image recognition, cognitive skills, commonsense reasoning, multi-step reasoning, image-based wordplay, etc., making this a challenging task for even current Vision-Language Models. In this paper, we present \rebus\ (Rebus Puzzle for the Word ``Rebus'', consisting of the ``Re'' - \faIcon{sync-alt} and ``Bus'' - \faIcon{bus} symbols), a large and diverse benchmark of $1,333$ English Rebus Puzzles containing different artistic styles and levels of difficulty, spread across 18 categories such as food, idioms, sports, finance, entertainment, etc. We also propose \rebusdescprogice, a model-agnostic framework which uses a combination of an unstructured description and code-based, structured reasoning, along with better, reasoning-based in-context example selection, improving the performance of Vision-Language Models on \rebus\ by $2.1-4.1\%$ and $20-30\%$ using closed-source and open-source models respectively compared to Chain-of-Thought Reasoning\footnote{The dataset and code are available at \url{https://github.com/abhi1nandy2/Re-Bus}}.
\end{abstract}




\maketitle

\section{Introduction}

\begin{figure}[h]
    \centering
    \includegraphics[width=0.7\linewidth]{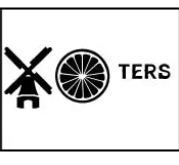}
    \caption{Example of a Rebus Puzzle in \rebus}
    \label{fig:rebus-example}
\end{figure}

Rebus Puzzles are a form of wordplay that uses images, letters, and symbols to represent words or syllables. They serve as a creative tool to spark lateral thinking, challenge conventional patterns, and invite solvers to uncover hidden meanings through visual clues. Understanding such puzzles requires 
a plethora of capabilities such as image recognition, commonsense knowledge and reasoning, multi-step reasoning, and understanding the creator's intent \cite{gritsevskiy2024rebus}. Fig. \ref{fig:rebus-example} shows an example of a Rebus Puzzle containing images and letters. The images in the puzzle are that of a ``Mill'' and ``Lime'', followed by letters that read ``Ters''. Combining ``Mill'', ``Lime'', and ``Ters'' creatively (by adding/subtracting/replacing letters), we get a meaningful word of ``Millimeters'' as the final answer of the puzzle. Note that the choice of the words for images are very crucial - for instance, if the word ``Turbine'' is used instead of ``Mill'', we would not arrive at a meaningful final answer.

Puzzle-solving and reasoning abilities of Vision-Language Models have been evaluated previously. For instance, M3Exam \cite{m3exam} evaluates multimodal multiple choice exam questions. MATH-Vision \cite{wang2024measuring} evaluates the mathematical reasoning ability of Vision-Language Models on math questions spanning several topics. There is also prior work on evaluating Rebus Puzzles in English \cite{gritsevskiy2024rebus} and in Italian \cite{italian-rebus} Languages. However, \textit{prior work on Rebus Puzzles neither proposes a diverse benchmark having different levels of difficulty, nor does any such work provide a model-agnostic solution that can be applied on top of both open as well as closed-source models with minimal or no training for improved performance}.

The development of Vision-Language Models (VLMs) \cite{bai2025qwen2, llava, gpt4, gemini, gpt-4o, agrawal2024pixtral} has witnessed a substantial rise in recent years. These models have demonstrated outstanding state-of-the-art (SOTA) performance across various downstream tasks, including Image Captioning and Visual Question Answering. These models are pre-trained such that images and text share a common embedding space, ensuring that images and their corresponding textual descriptions have similar representations within that space.

In this paper, we first inspect whether VLMs are able to understand and solve Rebus Puzzles - Given an image of a Rebus Puzzle (like in Fig. \ref{fig:rebus-example}), the VLM is expected to generate a natural language answer to the puzzle as a word/phrase (``Millimeters'' in case of Fig. \ref{fig:rebus-example}). This is a challenging task that extends beyond mere image analysis and linguistic comprehension, as it involves a layered process that draws on factual knowledge, contextual insight, language skills, and logical reasoning within specific boundaries—core abilities essential for tackling many real-world challenges \cite{italian-rebus}.

To evaluate the task, we curated a large and diverse multimodal dataset \rebus\ comprising of $1,333$ English Rebus Puzzles\footnote{The answer to every puzzle is in English.}, where each dataset sample contains an image of a Rebus Puzzle which contains a combination of images and/or texts, along with the answer to the puzzle, and a rich suite of meticulously annotated metadata such as a hint to solve the the puzzle, difficulty of the puzzle, whether the spelling of the text/objects in the image is varied in order to get the puzzle's answer, is color of text in the puzzle relevant in solving the puzzle, etc., making our proposed \rebus\ superior to that of the previous works on Rebus Puzzles \cite{gritsevskiy2024rebus, italian-rebus}. Also, to increase the diversity and difficulty of the puzzles, some samples in \rebus\ are generated using ControlNet \cite{controlnet}, which adds an ambient background as a backdrop while preserving the core content of the Rebus puzzle. This added background serves as a visual distraction, thereby increasing the difficulty of solving the puzzle. 

Additionally, we propose a compute-efficient \rebusdescprogice\ framework, which combines structured (code-based) and unstructured reasoning in an in-context learning setup, along with a lightweight example selection strategy requiring only minimal training. Unlike baselines that rely on a single reasoning style, \rebusdescprogice\ consistently improves puzzle-solving performance. For GPT-4o, it yields steady gains (Word-Level F1 rising from 0.489 in zero-shot normal prompting to 0.512 in three-shot \rebusdescprogice). The benefits are more striking for open-source models: Qwen2-VL-7B achieves up to a 20–30\% relative improvement over description-only and VisProg baselines (e.g., from 0.2 to 0.264 F1). These results highlight that the synergy of structured and unstructured reasoning, coupled with informed example selection, is key to unlocking better performance, particularly for weaker open-source VLMs.

We make the following contributions in this paper - (1) We introduce \rebus, a large, diverse dataset of $1,333$ annotated English Rebus Puzzles (2) We enhance puzzle difficulty using ControlNet to add distracting yet realistic visual backgrounds (3) We propose \rebusdescprogice, a compute-efficient framework combining structured and unstructured reasoning in-context (4) We design a novel in-context example selection strategy aligned with anticipated VLM reasoning patterns.

\section{Background}

\textbf{Linguistic Puzzle Solving. }Linguistic Puzzles have emerged as an intriguing benchmark for evaluating the reasoning and language capabilities of large language models (LLMs) \cite{rozner2021decrypting, manna2024riddle, giadikiaroglou2024puzzle}. While early research predominantly explored English-language challenges like crosswords \cite{wallace2022automated, littman2002probabilistic, ernandes2005webcrow, sadallah2024llms}, recent efforts have expanded to include a richer variety of puzzle types, including popular games such as Wordle \cite{anderson2022finding} and the New York Times Connections \cite{todd2024missed}. Beyond English, automated puzzle solvers have been developed for other languages as well—such as French \cite{angelini2023webcrow}, German \cite{zugarini2023ratselrevolution}, and Italian \cite{angelini2005solving, zugarini2024clue}. Additionally, educational puzzle generators are available in languages like Italian \cite{zeinalipour2023italian} and Turkish \cite{zeinalipour2024turkish}, highlighting the growing global interest in computational approaches to linguistic games. 

\noindent \textbf{Code-based reasoning using VLMs and LLMs. }Structured, code-based reasoning shows improvements in performance in complex, commonsense reasoning tasks. Recent approaches like VISPROG \cite{visprog} extend this paradigm to vision-language tasks, where VLMs and LLMs collaborate by generating modular code that orchestrates vision models and logical operations to solve complex visual problems. Code-based reasoning methods like PoT (Program of Thoughts) \cite{chen2023program} help LLMs tackle math by writing and running code, separating thought from calculation. PAL (Program-Aided Language Models) \cite{gao2023pal} boost LLM performance by turning text problems into code, allowing a Python Interpreter handle the computation. \citet{madaan-etal-2022-language} show that even without code in the task, code LLMs do better when commonsense problems are framed as code generation.    

\section{\rebus\ Dataset}

\subsection{Our Annotation Pipeline}

\begin{figure*}[t]
    \centering
    \includegraphics[width=0.9\linewidth]{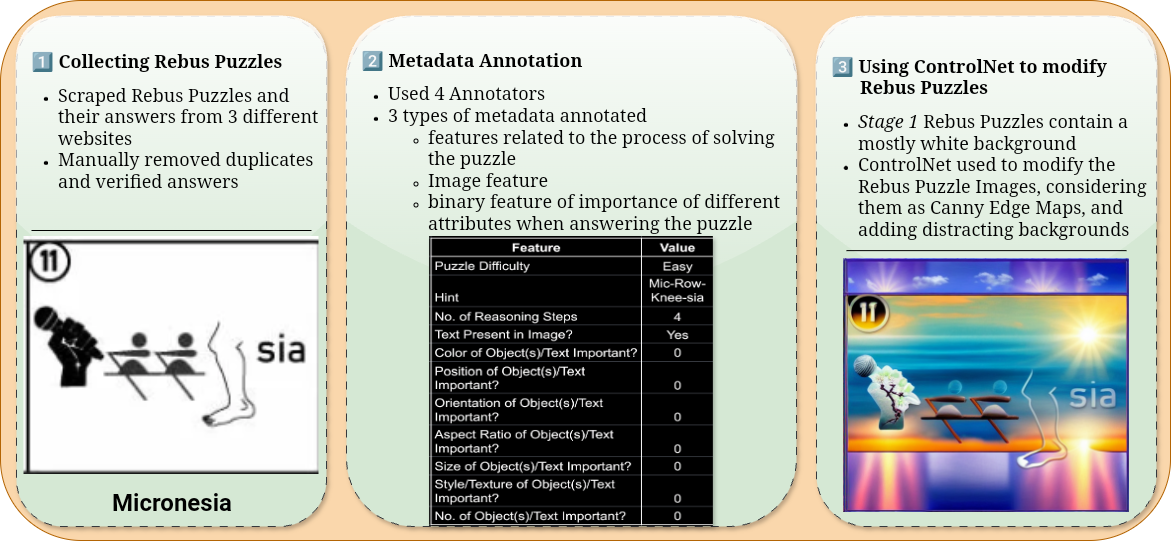}
    \caption{Annotation Pipeline of \rebus\ Dataset}
    \label{fig:rebus-annot}
\end{figure*}

The entire data collection and annotation pipeline is shown in Fig. \ref{fig:rebus-annot}. We curated a collection of Rebus Puzzles with meticulously annotated  metadata in this section in 3 stages.

\subsubsection{\textbf{\textit{Stage 1}: Collecting Rebus Puzzles from multiple Internet Sources}}
\label{sec:image_collection}

We collect Rebus Puzzle Images along with the corresponding ground truth answers from 3 different sources  - \url{https://eslvault.com/free-printable-rebus-puzzles/} (contains a diverse set of rebus puzzles that are mostly in black and white), \url{https://kids.niehs.nih.gov/games/brainteasers/rebus-puzzles} (contains mostly text-based rebus puzzles), and \url{http://flashbynight.com/rebus} (contains a diverse set of rebus puzzles that are mostly colored). After removing duplicate Rebus Puzzle Images, we end up with $722$ Rebus Puzzles. We also verified the answers collected for each Rebus Puzzle and made manual corrections to the answer wherever necessary. 

\subsubsection{\textbf{\textit{Stage 2}: Annotation of Rebus Puzzle Metadata}}
\label{sec:annotation}

\begin{figure}[H]
    \centering
    \includegraphics[width=0.85\linewidth]{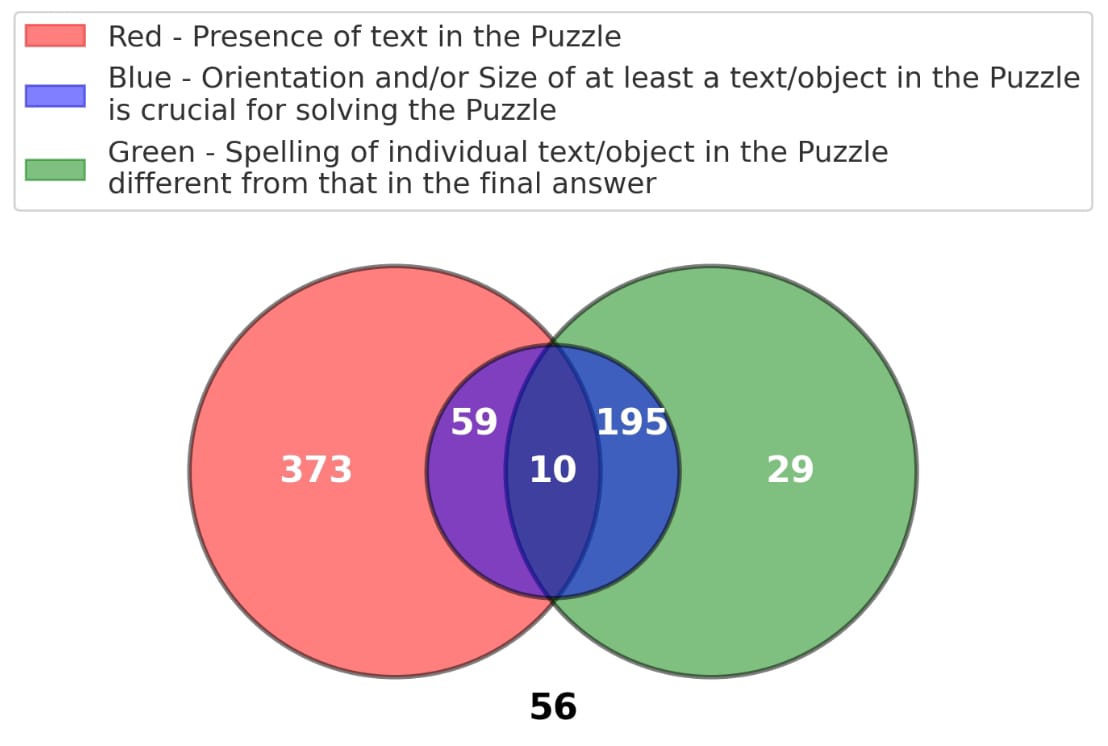}
    \caption{Breakdown of some important Rebus Puzzle Metadata Characteristics}
    \label{fig:feats-venn}
\end{figure}

Several types of metadata are annotated for the rebus Puzzles using $4$ annotators, all of whom were at least in their second year of undergraduate study and enrolled in institutions where English is the primary language of instruction. The annotated metadata includes - (1) features related to the process of solving the puzzle, such as puzzle difficulty (Easy/Hard), whether spelling of individual objects/text is different from that in the ground truth answer, hint for solving it, number of units of reasoning to solve it (2) image feature, like whether any text is present in the image (3) binary feature of importance of different attributes such as color, position, orientation, aspect ratio, size, style/texture, and number of object(s)/text when answering the puzzle. Fig. \ref{fig:feats-venn} shows the distribution of the puzzles across 3 types of binary metadata as a Venn  Diagram. This shows that Rebus Puzzles are highly diverse, as they are spread out across different combinations of the binary values of the metadata.   

\subsubsection{\textbf{\textit{Stage 3}: Using ControlNet to obtain modified versions of Rebus Puzzles}}

The Rebus Puzzles collected in \textit{Stage 1} contain a mostly white background, making the puzzle potentially easier to solve. One way to make a rebus Puzzle difficult to solve could be to add a distracting background to the Rebus Puzzle Image, without affecting the Rebus Puzzle in itself. To do so, we use ControlNet \cite{controlnet} on all the Rebus Puzzles collected in \textit{Stage 1}, treating the puzzles as Canny Edge Maps \footnote{API Calls to \url{https://huggingface.co/spaces/hysts/ControlNet} were performed during implementation}. These generated images are verified by a qualified annotator to keep only those images in the dataset that are meaningful and would have the same answer as the original puzzle (from \textit{Stage 1)}. Among the $722$ generated images, $611$ puzzle images are filtered, and are added to the \rebus\ dataset (along with the \textit{Stage 1} puzzles), making the total number of Rebus Puzzles  in \rebus\ Dataset as $1,333$. 

\subsection{Dataset Description and Details}

\begin{figure}[H]
    \centering
    \includegraphics[width=\linewidth]{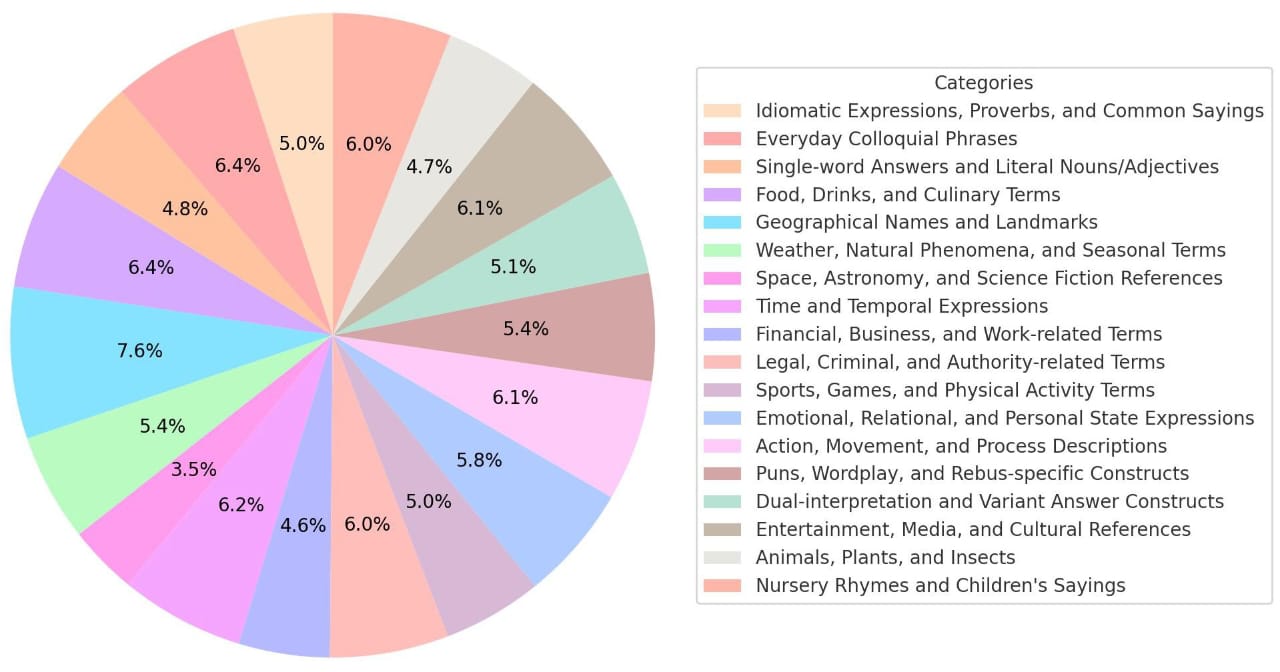}
    \caption{Distribution of Rebus Puzzles based on their category}
    \label{fig:categories-dist}
\end{figure}

The \rebus\ dataset has a total of $1,333$ images of Rebus Puzzles, $611$ of which are generated using ControlNet \cite{controlnet} and are therefore of a different artistic style.

The Rebus puzzles included in the \rebus\ Dataset encompass a diverse range of linguistic and conceptual features. To better understand these underlying patterns, we employ ChatGPT \cite{gpt-4o} to systematically categorize the ground truth answers into distinct and meaningful classes by designing an appropriate prompting strategy. Fig. \ref{fig:categories-dist} shows the distribution of the Rebus Puzzles across the 18 fine-grained categories predicted by ChatGPT. These categories belong to varied aspects of Language and Expression Usage, Knowledge and Facts, Culture and Society, Activities and Hobbies, and Nature and Living Things.

\begin{figure}[H]
    \centering
    \includegraphics[width=\linewidth]{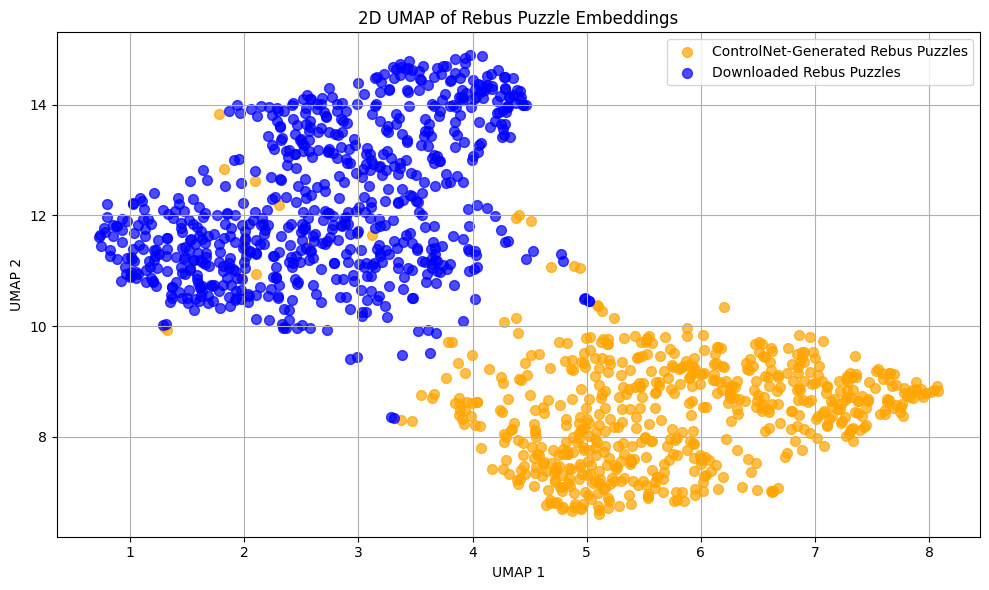}
    \caption{2D UMAP Representations of CLIP Image representations of Rebus Puzzle Images}
    \label{fig:image_embs}
\end{figure}

To illustrate the diversity among these Rebus Puzzle Images, we project their pre-trained CLIP \cite{clip} (MIT License) image embeddings into a 2D space using UMAP \cite{umap}, as shown in Fig. \ref{fig:image_embs}. We color-code the image samples according to their artistic styles. Interestingly, despite sharing identical puzzle answers, the Rebus Puzzle images generated via ControlNet \cite{controlnet} are semantically far apart from their original counterparts. This results in a highly diverse set of Rebus Puzzle images that span multiple categories.

\subsection{Proposed approach: \rebusdescprogice}

Our proposed approach \rebusdescprogice\ introduces a novel LLM-agnostic reasoning module in an in-context learning setting. The reasoning to be generated contains  2 components - (1) an \textbf{unstructured and detailed image description} (2) a \textbf{structured, code-based reasoning} component which elaborates the steps to be followed create the Rebus Puzzle Image. The unstructured and code-based reasoning components provide explicit factual knowledge and procedural logic, which are necessary in order to solve a Rebus Puzzle correctly. Also, we use a novel in-context example selection method based on the similarity between the code-based reasoning components (similar to \citet{poesia2022synchromesh}).

\noindent \textbf{In-Context Example Selection in \rebusdescprogice. }In our approach, a VLM is guided using task-specific examples and instructions provided within the same session, without modifying its underlying parameters. This setup enhances the VLM's output by leveraging relevant contextual information. To ensure the selected in-context examples are useful for the target task, we employ a vector embedding-based similarity technique to retrieve training examples whose embeddings closely match that of the test input. Additionally, we propose a novel technique (inspired by \citet{poesia2022synchromesh}) to learn a unified embedding that effectively represents a Rebus Puzzle Image.

\section{Experiments and Results}

\subsection{Experimental Setup}

All experiments using open-source models are carried out on 4 L40 GPUs, each having a VRAM of 48GB.

\subsection{Baselines}

We use the following prompting strategies as prompts - (1) \textbf{Zero-Shot. }This follows the naive zero-shot prompting strategy inspired by \citet{gritsevskiy2024rebus} (2) \textbf{Few-Shot CoT (Chain-of-Thought) \cite{cot}. }In addition to mentioning the task instructions, this baseline uses In-Context Learning \cite{icl}, where each in-context example contains the Rebus Puzzle image and a corresponding hint (annotated as metadata) as the input, and the corresponding ground truth answer as the output (3) \textbf{Few-shot with Descriptions. }This uses a similar prompt template as the previous baseline, where instead of using an annotated hint, an image description generated using GPT-4o \cite{gpt-4o} in a zero-shot setting is used as the intermediate reasoning in the in-context examples. Note that for the last two baselines, the in-context example(s) for each test sample are randomly sampled from examples in the holdout set. 

We benchmarked three closed-source models—GPT-4o, GPT-4o-mini, and GPT-4 Turbo—and three open-source vision-language models—Phi-3.5-Vision, Pixtral-12B, and Qwen2-VL-7B. GPT-4o stands out with its seamless support for text, audio, images, and video, achieving state-of-the-art multilingual, vision, and audio understanding while operating faster and more cost-efficiently than GPT-4 Turbo \cite{OpenAI2024GPT4oSystemCard,OpenAI2024GPT4oBenchmarks}. The lighter GPT-4o-mini, obtained through model distillation, preserves much of GPT-4o’s multimodal capabilities at significantly lower cost and latency, excelling in reasoning and coding benchmarks \cite{OpenAI2024GPT4oMini,EverydayAI2024GPT4oMini}. GPT-4 Turbo similarly offers strong performance in text and code tasks but with comparatively less advanced multimodal integration \cite{Time2024GPT4Turbo}. On the open-source side, Phi-3.5-Vision is a lightweight, multimodal model adept at dense reasoning and efficient image processing, even enabling high-quality OCR and text extraction in resource-constrained environments \cite{Microsoft2025Phi35Vision,Dyland2024Phi35OCR}. Pixtral-12B, a 12-billion-parameter model, delivers strong instruction-following performance in both text and vision, outperforming larger open models in multimodal benchmarks thanks to its high-resolution vision encoder and long-context support \cite{agrawal2024pixtral}. Lastly, Qwen2-VL-7B introduces dynamic-resolution visual tokenization and multimodal rotary embeddings (M-ROPE), enhancing its ability to process variable-resolution images and fuse visual and textual information—reaching performance comparable to leading closed-source models in some benchmarks \cite{Wang2024Qwen2VL}. Each model thus presents a distinct balance between capability, modality support, and computational accessibility, offering a diverse testbed for evaluating rebus puzzle solving.

\subsection{Automated Evaluation Metrics}

For automated evaluation, we employ two lexical text-matching metrics. The \textbf{word-level F1 score} is computed as the harmonic mean of precision and recall over the tokenized prediction and reference answers, providing a balanced measure of both answer completeness and correctness \cite{rajpurkar-etal-2016-squad}. In addition, we report \textbf{substring accuracy}, which measures whether the predicted answer occurs as a contiguous substring within the reference. This metric is particularly relevant for ReBus puzzle-solving, where reference answers may permit multiple surface realizations, and a prediction that partially overlaps with the ground truth can still capture essential semantic content.

\subsection{\rebusdescprogice\ vs. Baselines}

\begin{table*}[t]
\centering
\caption{Substring Accuracy across models with varying number of in-context examples and prompting methods.}
\label{tab:substring}
\begin{adjustbox}{max width=\textwidth}
\begin{tabular}{lccccccccccccccccc}
\toprule
\multirow{2}{*}{Model} & \multicolumn{2}{c}{Zero} & \multicolumn{5}{c}{One} & \multicolumn{5}{c}{Two} & \multicolumn{5}{c}{Three} \\
\cmidrule(lr){2-3} \cmidrule(lr){4-8} \cmidrule(lr){9-13} \cmidrule(lr){14-18}
& normal & cot & normal & cot & VisProg & only desc & \rebusdescprogice & normal & cot & VisProg & only desc & \rebusdescprogice & normal & cot & VisProg & only desc & \rebusdescprogice \\
\midrule
GPT-4o & 0.420 & 0.449 & 0.416 & 0.426 & 0.387 & 0.402 & 0.414 & 0.414 & 0.423 & 0.380 & 0.428 & 0.411 & 0.416 & 0.415 & 0.383 & 0.434 & 0.422 \\
GPT-4o-mini & 0.213 & 0.428 & 0.223 & 0.215 & 0.224 & 0.230 & 0.208 & 0.211 & 0.238 & 0.221 & 0.227 & 0.230 & 0.223 & 0.202 & 0.229 & 0.188 & 0.208 \\
GPT-4 turbo & 0.279 & 0.410 & 0.346 & 0.319 & 0.315 & 0.303 & 0.324 & 0.342 & 0.327 & 0.307 & 0.321 & 0.328 & 0.342 & 0.319 & 0.328 & 0.313 & 0.343 \\
Phi-3.5 & 0.169 & 0.114 & 0.075 & 0.163 & 0.128 & 0.086 & 0.110 & 0.066 & 0.152 & 0.099 & 0.071 & 0.098 & 0.062 & 0.154 & 0.087 & 0.062 & 0.096 \\
Pixtral & 0.107 & 0.096 & 0.102 & 0.065 & 0.075 & 0.108 & 0.096 & 0.086 & 0.066 & 0.075 & 0.093 & 0.078 & 0.095 & 0.083 & 0.093 & 0.092 & 0.093 \\
Qwen2-VL-7B & 0.342 & 0.161 & 0.268 & 0.209 & 0.208 & 0.160 & 0.101 & 0.241 & 0.182 & 0.185 & 0.139 & 0.068 & 0.343 & 0.146 & 0.214 & 0.111 & 0.107 \\
\bottomrule
\end{tabular}
\end{adjustbox}
\end{table*}

\begin{table*}[t]
\centering
\caption{Word-Level F1 Score across models with varying number of in-context examples and prompting methods}
\label{tab:f1}
\begin{adjustbox}{max width=\textwidth}
\begin{tabular}{lccccccccccccccccc}
\toprule
\multirow{2}{*}{Model} & \multicolumn{2}{c}{Zero} & \multicolumn{5}{c}{One} & \multicolumn{5}{c}{Two} & \multicolumn{5}{c}{Three} \\
\cmidrule(lr){2-3} \cmidrule(lr){4-8} \cmidrule(lr){9-13} \cmidrule(lr){14-18}
& normal & cot & normal & cot & VisProg & onlydesc & \rebusdescprogice & normal & cot & VisProg & only desc & \rebusdescprogice & normal & cot & VisProg & only desc & \rebusdescprogice \\
\midrule
GPT-4o & 0.489 & 0.467 & 0.503 & 0.516 & 0.507 & 0.514 & 0.513 & 0.511 & 0.549 & 0.506 & 0.521 & 0.517 & 0.519 & 0.517 & 0.506 & 0.536 & 0.512 \\
GPT-4o-mini & 0.330 & 0.457 & 0.355 & 0.356 & 0.355 & 0.346 & 0.352 & 0.358 & 0.370 & 0.362 & 0.374 & 0.366 & 0.348 & 0.361 & 0.366 & 0.360 & 0.352 \\
GPT-4 turbo & 0.382 & 0.451 & 0.433 & 0.421 & 0.424 & 0.413 & 0.398 & 0.440 & 0.432 & 0.423 & 0.426 & 0.439 & 0.442 & 0.438 & 0.426 & 0.422 & 0.431 \\
Phi-3.5 & 0.153 & 0.130 & 0.093 & 0.161 & 0.130 & 0.191 & 0.198 & 0.112 & 0.178 & 0.177 & 0.196 & 0.205 & 0.106 & 0.198 & 0.172 & 0.196 & 0.177 \\
Pixtral & 0.151 & 0.185 & 0.161 & 0.189 & 0.216 & 0.209 & 0.209 & 0.170 & 0.207 & 0.214 & 0.199 & 0.225 & 0.180 & 0.209 & 0.238 & 0.213 & 0.239 \\
Qwen2-VL-7B & 0.179 & 0.120 & 0.176 & 0.206 & 0.219 & 0.200 & 0.241 & 0.211 & 0.230 & 0.235 & 0.222 & 0.270 & 0.264 & 0.237 & 0.248 & 0.250 & 0.264 \\
\bottomrule
\end{tabular}
\end{adjustbox}
\end{table*}

Tables \ref{tab:substring} and \ref{tab:f1} show the substring accuracy and word-level F1 scores respectively across closed and open-source models with varying number of in-context examples and prompting methods. We can infer that - (1) The closed-source models (GPT-4o, GPT-4o-mini, GPT-4 turbo) consistently outperform open-source models (Phi-3.5, Pixtral, Qwen2-VL-7B) across both metrics. For instance, in Table 2 (Word-Level F1 Score), GPT-4o reaches 0.536 (three-shot, only description), while open-source models peak around 0.270 (Qwen2-VL-7B, three-shot, \textsc{RebusDescProgICE}). This highlights the superior reasoning and alignment capabilities of closed-source VLMs, particularly GPT-4 variants, which show more robust performance across prompting strategies. (2) Our method \rebusdescprogice\ shows notable improvements, particularly when compared to simpler prompting methods like "only description." For example, in GPT-4o (Table \ref{tab:substring}), three-shot \rebusdescprogice\ achieves 0.422 substring accuracy, comparable to or better than most other settings. Similarly, in Table 2, GPT-4o reaches 0.512 F1, showing stable gains. Even for weaker open-source models like Qwen2-VL-7B, \rebusdescprogice\ boosts performance substantially (e.g., from 0.200 to 0.241 in one-shot F1, and up to 0.264 in three-shot F1), suggesting that the synergy of visual program + description generalizes across model families. (3) Increasing the number of in-context examples generally leads to modest but consistent improvements, especially in F1 scores. For example, GPT-4 turbo F1 improves from 0.382 (zero-shot normal) to 0.442 (three-shot normal). Substring accuracy shows smaller gains, but still some improvements (e.g., GPT-4o from 0.420 zero-shot normal to 0.416–0.422 three-shot variants). However, gains plateau beyond two or three examples, indicating diminishing returns. (4) Importance of combining VisProg and Description (our method).
Comparing isolated prompting methods highlights why both components are essential. VisProg alone (e.g., GPT-4o three-shot VisProg: 0.383 substring acc, 0.506 F1) or only description (GPT-4o three-shot: 0.434 substring acc, 0.536 F1) do well individually, but \textsc{RebusDescProgICE} consistently balances both to achieve competitive performance (0.422 substring acc, 0.512 F1). In open-source models, this effect is even clearer: for Qwen2-VL-7B, VisProg (0.214 substring acc, 0.248 F1) and only desc (0.111 substring acc, 0.250 F1) underperform compared to \textsc{RebusDescProgICE} (0.107 substring acc, 0.264 F1). This demonstrates that combining structured visual reasoning (VisProg) with descriptive context leads to more reliable gains than either alone.

\noindent \textbf{Performance on Augmented Test Data. }

\begin{table*}[t]
\centering
\caption{Substring Accuracy \textbf{on Augmented Test Data} across models with varying number of in-context examples and prompting methods.}
\label{tab:aug-substring}
\begin{adjustbox}{max width=\textwidth}
\begin{tabular}{lccccccccccccccc}
\toprule
\multirow{2}{*}{Model} & \multicolumn{5}{c}{One} & \multicolumn{5}{c}{Two} & \multicolumn{5}{c}{Three} \\
\cmidrule(lr){2-6} \cmidrule(lr){7-11} \cmidrule(lr){12-16}
& normal & cot & VisProg & only desc & \rebusdescprogice & normal & cot & VisProg & only desc & \rebusdescprogice & normal & cot & VisProg & only desc & \rebusdescprogice \\
\midrule
GPT-4o & 0.257 & 0.264 & 0.268 & 0.254 & 0.268 & 0.262 & 0.248 & 0.246 & 0.280 & 0.246 & 0.262 & 0.254 & 0.245 & 0.259 & 0.259 \\
Phi-3.5 & 0.060 & 0.118 & 0.093 & 0.053 & 0.058 & 0.079 & 0.120 & 0.081 & 0.039 & 0.033 & 0.079 & 0.104 & 0.079 & 0.025 & 0.039 \\
Pixtral & 0.065 & 0.074 & 0.065 & 0.074 & 0.062 & 0.079 & 0.041 & 0.060 & 0.046 & 0.058 & 0.086 & 0.058 & 0.072 & 0.067 & 0.048 \\
Qwen2-VL-7B-Instruct & 0.276 & 0.282 & 0.123 & 0.116 & 0.097 & 0.238 & 0.165 & 0.107 & 0.083 & 0.083 & 0.222 & 0.129 & 0.090 & 0.067 & 0.070 \\
\bottomrule
\end{tabular}
\end{adjustbox}
\end{table*}

\begin{table*}[t]
\centering
\caption{Word-Level F1 score \textbf{on Augmented Test Data} across models with varying number of in-context examples and prompting methods.}
\label{tab:aug-wordf1}
\begin{adjustbox}{max width=\textwidth}
\begin{tabular}{lccccccccccccccc}
\toprule
\multirow{2}{*}{Model} & \multicolumn{5}{c}{One} & \multicolumn{5}{c}{Two} & \multicolumn{5}{c}{Three} \\
\cmidrule(lr){2-6} \cmidrule(lr){7-11} \cmidrule(lr){12-16}
& normal & cot & VisProg & only desc & \rebusdescprogice & normal & cot & VisProg & only desc & \rebusdescprogice & normal & cot & VisProg & only desc & \rebusdescprogice \\
\midrule
GPT-4o & 0.366 & 0.400 & 0.379 & 0.372 & 0.383 & 0.384 & 0.395 & 0.397 & 0.395 & 0.374 & 0.381 & 0.385 & 0.402 & 0.400 & 0.391 \\
Phi-3.5 & 0.085 & 0.134 & 0.135 & 0.152 & 0.164 & 0.125 & 0.161 & 0.169 & 0.163 & 0.164 & 0.125 & 0.173 & 0.154 & 0.156 & 0.158 \\
Pixtral & 0.149 & 0.186 & 0.188 & 0.192 & 0.195 & 0.150 & 0.190 & 0.190 & 0.174 & 0.201 & 0.161 & 0.209 & 0.206 & 0.195 & 0.215 \\
Qwen2-VL-7B-Instruct & 0.209 & 0.237 & 0.185 & 0.183 & 0.218 & 0.212 & 0.234 & 0.207 & 0.207 & 0.235 & 0.213 & 0.233 & 0.208 & 0.212 & 0.253 \\
\bottomrule
\end{tabular}
\end{adjustbox}
\end{table*}

Tables \ref{tab:aug-substring} and \ref{tab:aug-wordf1} display the Substring Accuracy and word-level F1 scores on Augmented Test Data with varying number of in-context examples and prompting
methods. We can infer that - (1) The overall low scores across models arise from the complexity of our dataset, where solving Rebus puzzles requires layered semantic reasoning; this is further amplified by the ControlNet-augmented noisy backgrounds, resulting in best scores remaining modest (e.g., maximum F1 of only 0.402 for GPT-4o). (2) GPT-4o consistently achieves the highest performance across substring accuracy (0.280 with two in-context examples, only desc) and word-level F1 (0.402 with two in-context examples, VisProg), reaffirming the challenging nature of the dataset even for state-of-the-art closed-source models. (3) Open-source models such as Phi-3.5 and Pixtral struggle considerably, with word-level F1 mostly below 0.20, reflecting their limited capacity for abstract reasoning under noisy conditions, whereas Qwen2-VL-7B-Instruct shows relatively better resilience (F1 up to 0.253). (4) Our proposed \rebusdescprogice\ framework provides consistent gains over standard prompting, particularly for weaker models — for instance, boosting Pixtral’s F1 from 0.186 (cot, one example) to 0.201 (two examples, \rebusdescprogice). (5) Increasing the number of in-context examples does not uniformly improve performance, confirming that puzzle-solving is not driven by pattern-matching; instead, structured reasoning guidance through \rebusdescprogice\ is crucial for robustness. Overall, these results establish our dataset as a strong benchmark for evaluating the reasoning capabilities of vision-language models under challenging, real-world-like conditions.


\section{Conclusion}

Closed-source VLMs remain far ahead of open-source ones in this challenging rebus puzzle-solving task. Nevertheless, our proposed method, \rebusdescprogice, proves robust across settings and especially beneficial for open-source models that otherwise struggle. While additional in-context examples help, the real performance boost comes from integrating both code-based reasoning (VisProg) and descriptive grounding, validating our design choice.



\bibliographystyle{ACM-Reference-Format}
\bibliography{sample-base}


\end{document}